\documentclass[twoside,twocolumn,10pt]{article}

\usepackage{wscg}           
\RequirePackage{ifpdf}
\ifpdf
 \RequirePackage[pdftex]{graphicx}
 \RequirePackage[pdftex]{color}
\else
 \RequirePackage[dvips,draft]{graphicx}
 \RequirePackage[dvips]{color}
\fi
\usepackage{booktabs}      

\usepackage{nopageno}       
\usepackage{lipsum}
\usepackage{graphicx}
\usepackage{times}
\usepackage{epsfig}
\usepackage{cite}
\usepackage{amsmath,amssymb,amsfonts}
\usepackage{algorithmic}
\usepackage{graphicx}
\usepackage{textcomp}
\usepackage{xcolor}
\usepackage{comment}
\usepackage{amsmath}
\usepackage{tabularx}
\usepackage[htt]{hyphenat}

\title{Calibration and Auto-Refinement for Light Field Cameras}

\author{
\parbox{0.6\textwidth}{\centering {Yuriy Anisimov}$^{1,2}$, {Gerd Reis}$^{1}$, {Didier Stricker}$^{1,2}$}\\*[2mm]
\parbox{0.3\textwidth}{\centering
$^{1}$German Research Center\\
for Artificial Intelligence,\\
Trippstadter Str. 122\\
67663 Kaiserslautern, Germany
}
\parbox{0.3\textwidth}{\centering
$^{2}$University of Kaiserslautern,\\
Gottlieb-Daimler-Str. 47\\
67663 Kaiserslautern, Germany
}
\\\\
\{firstname.lastname\}@dfki.de
}

\usepackage{url}
\makeatletter
\def\Uslash{\mathbin{\mathchar`\/}\@ifnextchar{/}{\kern-.15em}{}}
\g@addto@macro\UrlSpecials{\do \/ {\Uslash}}
\def\Ucolon{\mathbin{\mathchar`:}\@ifnextchar{/}{\kern-.1em}{}}
\g@addto@macro\UrlSpecials{\do : {\Ucolon}}
\makeatother

\def\eg{\emph{e.g. }} 
\def\ie{\emph{i.e. }}

\def\wrt{\emph{w.r.t. }} 
\def\etal{\emph{et al. }}

\newcommand{\Mod}[1]{\ \mathrm{mod}\ #1}

\begin{document}

\twocolumn[{\csname @twocolumnfalse\endcsname

\maketitle  

\begin{abstract}
\noindent
The ability to create an accurate three-dimensional reconstruction of a captured scene draws attention to the principles of light fields.
This paper presents an approach for light field camera calibration and rectification, based on pairwise pattern-based parameters extraction.
It is followed by a correspondence-based algorithm for camera parameters refinement from arbitrary scenes using the triangulation filter and nonlinear optimization.
The effectiveness of our approach is validated on both real and synthetic data.
\end{abstract}

\subsection*{Keywords}
light field, camera calibration, camera rectification, calibration refinement

\vspace*{1.0\baselineskip}
}]


\copyrightspace
\section{Introduction} \label{introduction}
Light field cameras \cite{ng2005light} utilize a multi-view principle, focusing incoming light on the image sensor over a grid of lenses. 
It creates a set of proportionally shifted images that can be used to reconstruct a three-dimensional representation of the captured scene. 
These are two types of light field cameras. 
As proposed by Adelson and Wang in \cite{adelson1992single}, a light field camera can be created from an array of micro-lenses placed in front of the camera sensor. 
An alternative way of constructing the light field camera was presented by Wilburn \etal in \cite{wilburn2005high} and involves placing an array of ordinary cameras with predetermined proportional distances between them. 
A different principle, which can be described as intermediate between the previous two, is represented in \cite{anisimov2019a}. 
There, the light field camera is built on a single camera sensor with an array of full-size lenses placed in front of it. 
Among the various applications of light fields, the estimation of depth maps attracts particular attention, as the large number of viewpoints increases the quality of the reconstruction and the simple geometry of the light fields simplifies the calculations.
Due to the inaccuracies in the placement of light field camera components (micro-lens array or single-view cameras) the direct estimation of scene depth information will not be accurate.
Natural lens distortion also affects the quality of the result. 
There are two ways to solve these problems.
For the reconstruction algorithm, either the exact lens positions and geometry should be taken into consideration, or the images from such a camera can be rectified to assure constraints of images' common rotation and proportional placement. 
One of the simplifications in reconstruction from multi-view systems has to do with limiting the search for matching correspondences to specific scan lines instead of searching the entire image.
Therefore, the rectification-based way of lens placement compensation should be used in the multi-view systems if their view placement allows the rectification without vanishing of significant image plane parts.
Camera calibration, or camera resectioning \cite{hartley2003multiple}, is a process of retrieving camera intrinsic and extrinsic parameters. 
Usually, the intrinsic parameters include the focal length, coordinates of a principal point, axis skew, and distortion coefficients. 
Extrinsic parameters contain information about the pose of the view in the form of rotation and translation \wrt the scene origin.
These parameters may change slightly under the mechanical or thermal influences during the operation of the light field cameras. 
In general, the light field camera must be recalibrated to cope with such changes, but sometimes this is not possible due to camera operating limitations or conditions. 
In such cases, the calibration data can be checked and corrected using arbitrary scene information.
We present an approach for light field camera calibration and rectification, which is based on a well-known pattern-based method from Zhengyou Zhang \cite{zhang2000flexible}. 
The results of the calibration are used to generate look-up tables that are applied to the light field views for their rectification.
The key feature of the method is its simplicity of implementation since this calibration requires a checkerboard pattern that can be produced on any conventional printer. 
At the same time, designed a nonlinear optimization model for estimating the compensation of changes in the position of the light field camera lenses. 
This method uses features extracted from an arbitrary scene.
To improve their accuracy, we introduce a triangulation-based correspondences filtering method, as well as a chaining method for tracking them in different light field views. 
Our algorithms were verified on a light field camera with single full-size lenses in an array.
It can be potentially used in micro-lens-based cameras, however, the potential misplacement of lenses in a manufactured array is not as large as in cameras with full-size lenses.
The paper is structured as follows. 
An overview of previous work is given in Section \ref{related_work}, Sections \ref{calibration} and \ref{refinement} outline the proposed methods, and experimental results for them are presented in Section \ref{experiments}, followed by a conclusion in Section \ref{conclusion}.
To prevent the term "camera" from being unambiguous, we use it in reference to a light-field camera; and each image taken from a particular lens is called a "view".

\section{Related work} \label{related_work}
The first paper describing the principles of light field calibration for camera arrays was proposed by Vaish \etal in \cite{vaish2004using}.
Using planar parallax, the authors retrieve the relative positions of the light field views and use them for further processing. 
Extension of stereo calibration and rectification principles for multi-view cameras is presented in the work of Kang \etal \cite{kang2008efficient}.
In our work similar principles were used for defining the common values of rectification parameters.
Approach of Xu \etal \cite{xu2015camera} optimizes all parameters of light field camera simultaneously by constrained bundle adjustment model.
Most of the approaches for light field calibration consider usage of micro-lens-based cameras.
They can't be applied directly to camera arrays due to the fact that the shift of the lenses relative to each other in such cameras is much higher than in the plenoptic cameras.
In the work of Bergamasco \etal \cite{bergamasco2015adopting} a parameter-free camera model is used for the light field rays representation for further triangulation-based calibration.
An approach of Jin \etal \cite{jin2016effective} estimates the centers of sub-images, aligns all views in a grid, and performs the rotation of all views to a common plane. 
Bok \etal in \cite{bok2016geometric} extracts lines from the raw-captured calibration pattern and attempts to compensate for lens distortion and misplacement inaccuracies using these features. 
Likewise, Noury \etal in \cite{noury2017light} employ the raw images; however, they use corners as the features for calibration procedure  along with nonlinear optimization of the result. 
Method of O'Brien \etal \cite{o2018calibrating} extracts disc features instead of conventional corners for retrieving the light field calibration data.
An approach by Sun \etal \cite{sun2019blind} works with three calibration targets at different distances together with gradient-based correspondences search. 
In the publication of Zhou \etal \cite{zhou2019light}, the calibration problem is solved by estimating the original homography solution using a calibration scheme and its subsequent nonlinear optimization.
Ji and Wu in \cite{ji2019calibration} propose a calibration model for plenoptic cameras calibration with a conventional calibration target.
\section{Calibration Algorithm} \label{calibration}
\begin{figure}[]
    \centering
    \includegraphics[width=7.5cm]{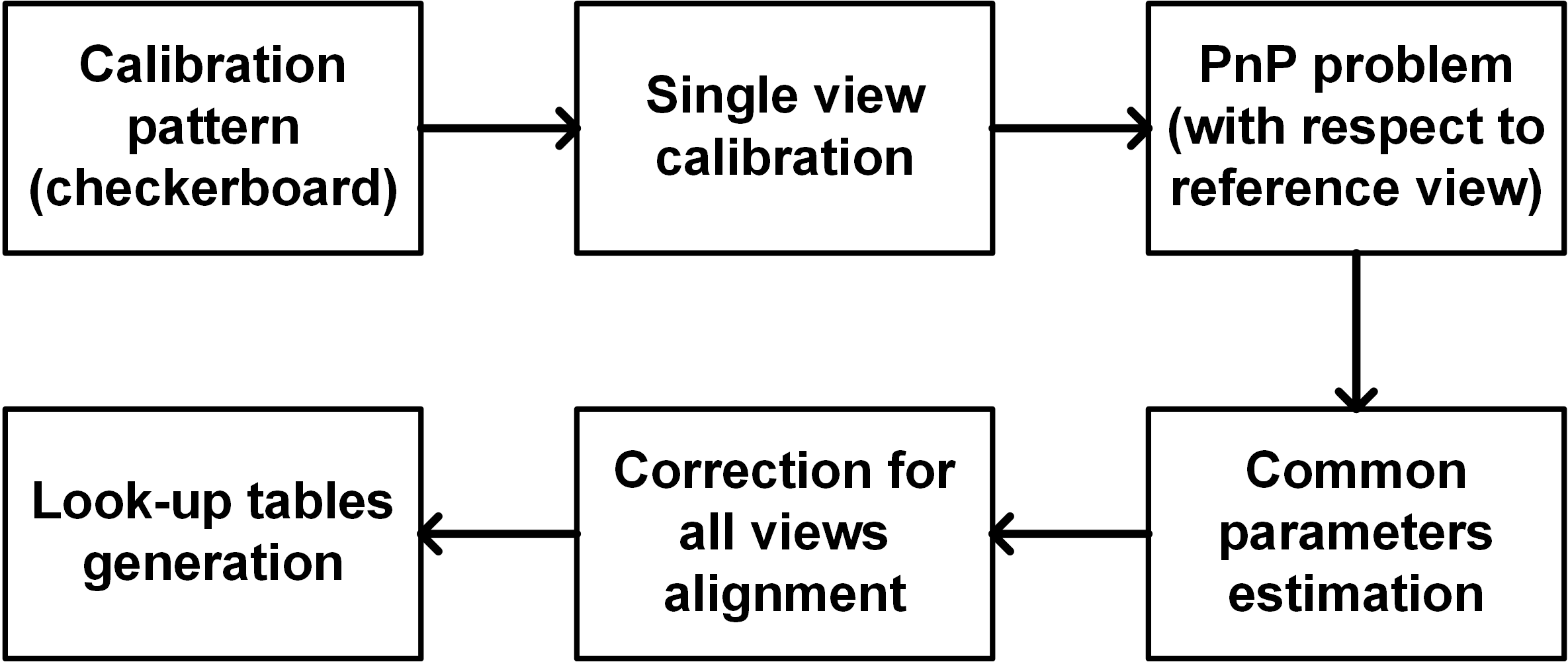}
    \caption{A pipeline of pattern-based calibration algorithm}
    \label{fig:pbc}
\end{figure}
A calibration algorithm estimates the intrinsic and extrinsic parameters of all light field views. 
This data is further used for the views rectification, which creates constrains for simplification of the depth estimation.
The pipeline of this method is presented in Fig. \ref{fig:pbc}.
\subsection{Camera model} \label{cm}
For a separated view of the light field a pinhole camera model is used \cite{hartley2003multiple}. 
The general relation of two-dimensional (2D) image points and three-dimensional (3D) scene points can be described as: 
\begin{align}
\label{eq:xy1}
\begin{split}
\begin{bmatrix}
x \\
y \\
1
\end{bmatrix}
= 
\begin{bmatrix}
f_x & s & c_x \\
0 & f_y & c_y \\
0 & 0 & 1
\end{bmatrix}
\begin{bmatrix}
r_{00} & r_{01} & r_{02} & t_0 \\
r_{10} & r_{11} & r_{12} & t_1\\
r_{20} & r_{21} & r_{22} & t_2
\end{bmatrix}
\begin{bmatrix}
X \\
Y \\
Z \\
1
\end{bmatrix}
\end{split}
\end{align}

\begin{align}
\label{eq:pkrt}
\begin{split}
P = K[R|t],\;m = PM 
\end{split}
\end{align}
where $m$ corresponds to homogeneous image point, $K$ -- camera calibration matrix, which consists of pixel focal lengths $f_{x}$ and $f_{y}$, principal point coordinates $c_{x}$ and $c_{y}$, and axis skew $s$. 
$R$ and $t$ correspond to rotation matrix and translation vector respectively. These components are assembled into a projection matrix $P$. $M$ stands for homogeneous world point. 
This representation does not take the lens distortions into account. 
Therefore, in our model we compensate the radial distortions up to their second order by \cite{hartley2003multiple}:
\begin{equation}
\label{eq:dist}
\begin{gathered}
x = x_{c} + (x_{n} - x_{c})(1+k_{1}r^{2}+k_{2}r^{4})\\
y = y_{c} + (y_{n} - y_{c})(1+k_{1}r^{2}+k_{2}r^{4})\\
r^{2} = (x_{n} - x_{c})^{2} + (y_{n} - y_{c})^{2}
\end{gathered}
\end{equation}
where $\{x; y\}$ are the undistorted points, $\{x_{n}; y_{n}\}$ correspond to original image points, $(x_{c}; y_{c})$ stands for radial distortion center, $r$ is the distance from this center to the distorted image point, and $k_{1}$ and $k_{2}$ are the radial distortion coefficients.
\begin{figure*}[]
    \centering
    \begin{tabular}{cc}
    \includegraphics[width=7cm]{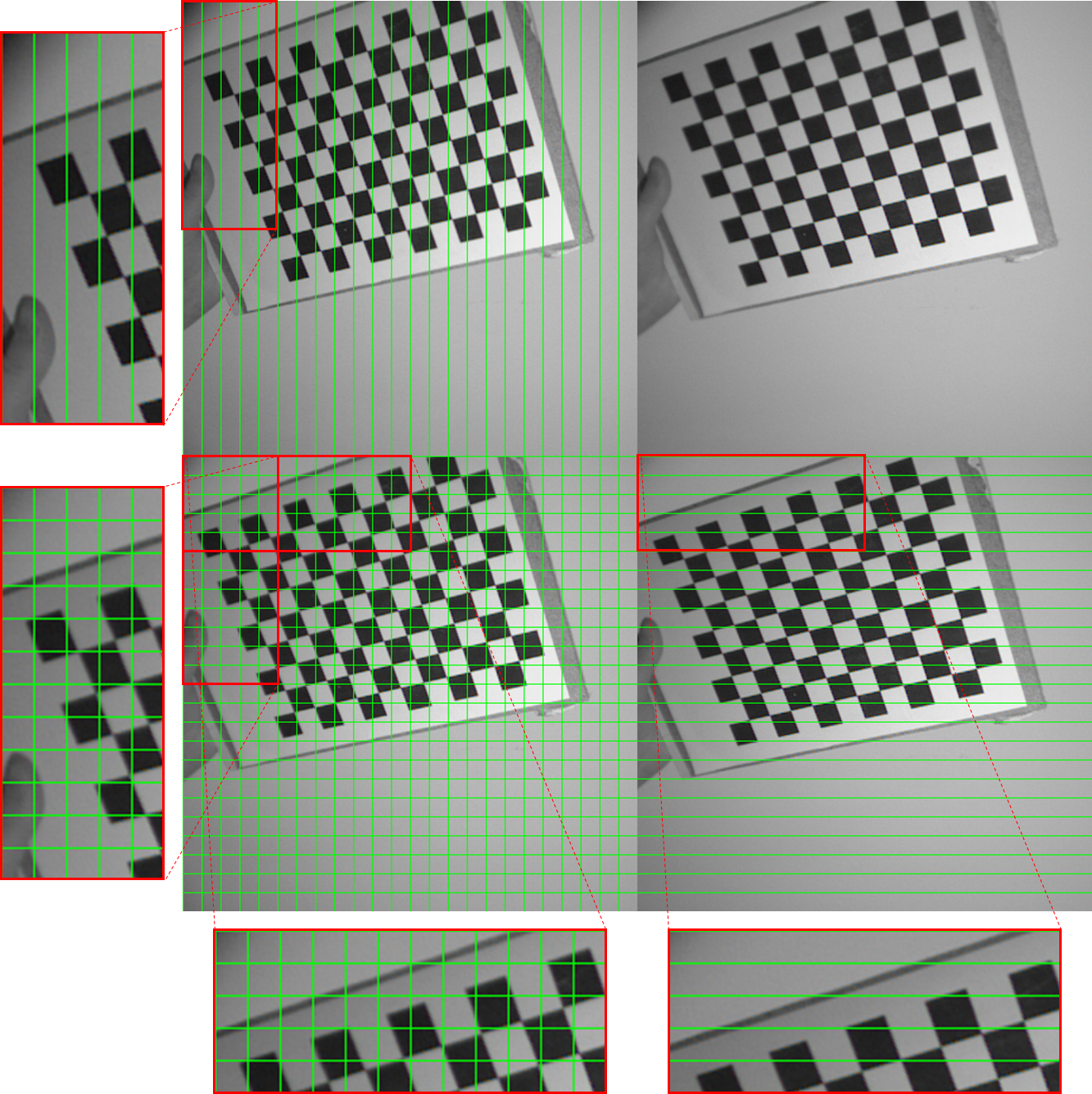} & \includegraphics[width=7cm]{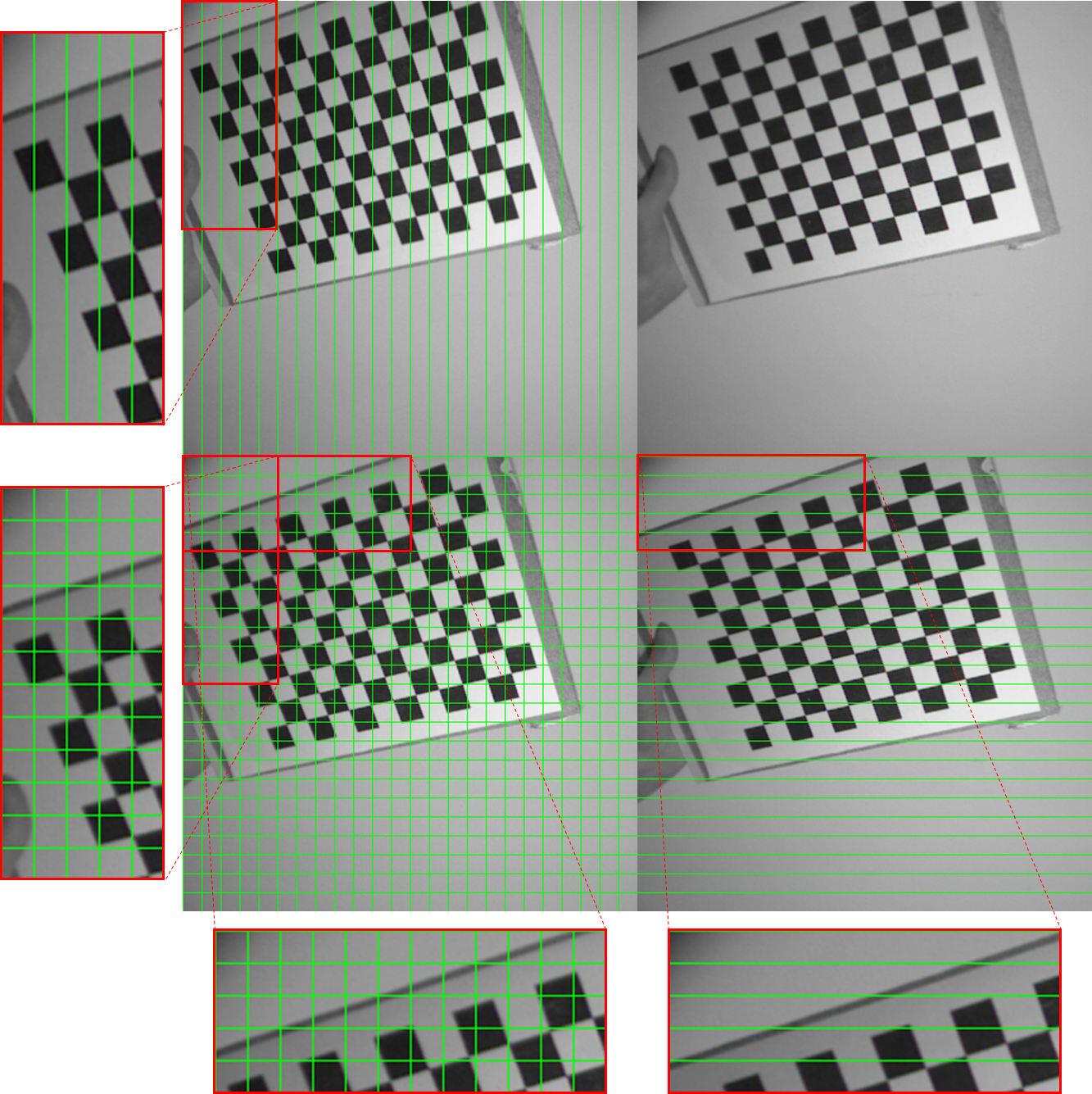} \\
    (a) & (b)
    \end{tabular}
    \caption{A subset of (a) original and (b) rectified light field after applying the algorithm from Section \ref{calibration}. The corners of the checkerboard are properly aligned, which confirms the rectification}
    \label{fig:before_after}
\end{figure*}
\subsection{Pattern-Based Calibration} \label{pbc}
A method, which uses a planar calibration pattern, was originally described by Zhengyou Zhang in \cite{zhang2000flexible}. 
The interest points of a calibration pattern should be such that they can be easily found by a corner detection algorithm.
A pattern with a known structure, such as a checkerboard, is used.
The measured physical size of the pattern elements is required for solving a task of view position determination with respect to the calibration target. 
It is used further to provide the correct values of the focal length, with which the pattern was captured.
First, the corners of the located checkerboard need to be found. 
Due to the high contrast of the black-and-white pattern, its elements can be easily segmented based on histogram analysis.
Corners' position is determined by the algorithm from \cite{suzuki1985topological}. 
A sub-pixel refinement step, described in \cite{foerstner1987a}, is used for obtaining accurate correspondences. 
This information is used to find the optimal intrinsic parameters of each view.
Having a set of points $\{m\}$, which size corresponds to the number of corners detected in the checkerboard, and its 3-dimensional matches $\{M\}$, common for all images, the algorithm computes the intrinsic matrix $K$ and the distortion coefficients vector $d$ in a way that these values reduce the reprojection error between the original and projected points with $K$ and $d$, assuming that images are taken with zero rotation and translation \wrt scene origin. 
By using Eq. \ref{eq:pkrt} with estimated intrinsic matrix $K$, and $R = I$, $t =  [0\;0\;0]$, where $I$ stands for identity matrix of size $3\times3$, the reprojections of 3D points can be found.
They are used afterward for estimating the root-mean-square error between originally detected points and the reprojected ones.
The 3D coordinates of pattern points $\{M\}$ are assumed to be known.
Their placement is defined by physical distances between real-world corners of the checkerboard, which can be expressed from the size of squares.
The algorithm results in the intrinsic values (focal length, camera center) of each light field view together with the distortion coefficient per lens. 
In order to reduce the reprojection error, the result is further optimized using the Levenberg-Marquardt algorithm \cite{marquardt1963algorithm}. 
It has been empirically established that this method requires dozens of images with different pattern positions to accurately estimate the calibration parameters.
The order of magnitude of this number can be explained by the need for the entire image space to be covered by the sum of the pattern positions among all captured frames. 
Results of single-camera calibration are further used for estimation of the relative parameters of every view \wrt the reference. 
During this step we fix extrinsic parameters of the reference view, so that $R_{ref} = I, t_{ref} = [0\;0\;0]$, and compute relative rotation and translation of every other view $\{R_{i}; t_{i}\}, i = 1..N$, where $N$ is a total number of views in the camera, based on their position in relation to scene center. 
It is possible since all views capture the same pattern position in every image. 
By having an assumption of known 3D points the Perspective-n-Point (PnP) problem \cite{fischler1981random} is solved for obtaining the rotation and translation of every view \wrt reference. 
As before, the results are subject to Levenberg-Marquardt optimization.
In the end, we obtain a set of original rotation and translation vectors $\{r^{O}, t^{O}\}$, which are used for the processing in Section \ref{refinement}.
Rotation vectors are obtained from rotations matrices $R^{O}$ by applying the Rodrigues transformation \cite{rodrigues1840on}.
It is used for the simplicity of calculations in the next steps.

\subsection{Multi-View Rectification}
The next step of the presented algorithm is the estimation of optimal values of the intrinsic and extrinsic parameters for all views.
It is necessary for the alignment of all views on the same image plane with proportional distances between every view.
First, all light field views should be rectified with a common intrinsic matrix $K_{r}$, which is  defined as the following:
\begin{equation}
\begin{gathered}
{f_{r}^{x}} = \frac{\sum_{i=1}^{N} f_{i}^{x}}{N}; {f_{r}^{y}} = \frac{\sum_{i=1}^{N} f_{i}^{y}}{N};\\
c_{r}^x = \frac{w}{2}; c_{r}^y = \frac{h}{2};\\
K_{r} = 
\begin{bmatrix}
{f_{r}^{x}} & 0 & c_{r}^x \\
0 & {f_{r}^{y}} & c_{r}^y \\
0 & 0 & 1
\end{bmatrix}
,
\end{gathered}
\label{eq:kr}
\end{equation}
where $N$ is the number of light field views, and $w, h$ correspond to width and height of every view. 
Rotation and translation vectors $\{r^{O}, t^{O}\}$ are used for getting an optimal value of common rotation and translation vector, to which all views would be brought by the rectification. 

Common rotation vector $r_r$ is estimated as an average of all rotation vectors as:
\begin{equation}
\begin{gathered}
r_{r} = \frac{\sum_{i=1}^{N} r^{O}_{i}}{N-1}
,
\end{gathered}
\end{equation}
and further converted to rotation matrix $R_{r}$ by applying Rodrigues transform. 
Computations of common translation vector involve the spatial information of every view.
Since the translation vectors are defined \wrt reference view, we estimate the relative values by involving $a$ and $b$ as vertical and horizontal spatial dimensions of light field, where $a \times b = N$. 
For a specific light field view  $i = 1..N$, the translation vector $t^V_i$ is formed from per-axis components as $t^{V}_i = [t^{V_x}_i\;t^{V_y}_i\;t^{V_z}_i]$ and estimated as a mean of all components not in the same row or column as: 
\begin{equation}
\begin{gathered}
\bar{a} = \lfloor{i/a}\rfloor; \bar{b} = i \Mod b;\\
t^{V_x}_i = 
\begin{cases}
t^{O_x}_i/(\hat{b} - \bar{b}) ,\hat{b} - \bar{b} \neq 0 \\
0 ,\hat{b} - \bar{b} = 0 
\end{cases}
\\
t^{V_y}_i = 
\begin{cases}
t^{O_y}_i/(\hat{a} - \bar{a}) ,\hat{a} - \bar{a} \neq 0 \\
0 ,\hat{a} - \bar{a} = 0 
\end{cases}
\\
t^{V_z}_i = 0\\
\end{gathered}
,
\end{equation}
where $\hat{a}$ and $\hat{b}$ stand for spatial coordinates of reference view, defined in Section \ref{pbc}.
All translation vectors form a set $t^V$. 
Common translation vector $\bar{t_{r}} = [{t_r^x}\;{t_r^y}\;0]$ is estimated as:
\begin{equation}
\begin{gathered}
S_{t^x} = \sum_{i=1}^{N}(1-\delta_{t^{V_x}_{i},0});
S_{t^y} = \sum_{i=1}^{N}(1-\delta_{t^{V_y}_{i},0})\\
{t_r^x} = \frac{\sum_{i=1}^{N} t^{V_x}_{i}}{S_{t^x}}; 
{t_r^y} = \frac{\sum_{i=1}^{N} t^{V_y}_{i}}{S_{t^y}}
\end{gathered}
,
\end{equation}
where $S_{t_x}$ and $S_{t_y}$ are numbers of particular non-zero components in $t^V$, found using Kronecker delta:
\begin{equation}
\delta_{x,y} = 
\begin{cases}
0, x \neq y \\
1, x = y
\end{cases}
.
\end{equation}
This vector is a basis for the set of per-view translation vectors $t^p$:
\begin{equation}
\begin{gathered}
t^{p_x}_i = t_r^x (\hat{b} - \bar{b});
t^{p_y}_i = t_r^y (\hat{a} - \bar{a});
t^{p_z}_i = 0
\end{gathered}
\label{eq:tr}
.
\end{equation}
Using the results from Eq. \ref{eq:kr}--\ref{eq:tr} we estimate the rectified projection matrix $P_{r}$ per every light field view $i = 1..N$ as \cite{hartley2003multiple}: 
\begin{equation}
\begin{gathered}
\bar{R_{i}} = R_{r} {R^{O}_{i}}^{T}\\
\bar{t_{i}} = \bar{R_{i}} t^{p}_{i}\\
P^{r}_{i} = K_{r}[\bar{R_{i}}|\bar{t_{i}}]
\end{gathered}
\end{equation}
The resulting camera projection matrix represents the ideal position of its corresponding light field view. 
The further remapping of pixels to the proper position is done by applying a look-up table, which stores per-pixel coordinates of the rectified image and generated using $P^{r}_{i}$, distortion coefficients and original intrinsic and extrinsic values.
Fig. \ref{fig:before_after} shows, how light field images changes after applying the described pipeline.

\section{Calibration Auto-Refinement} \label{refinement}
\begin{figure}[]
    \centering
    \includegraphics[width=7cm]{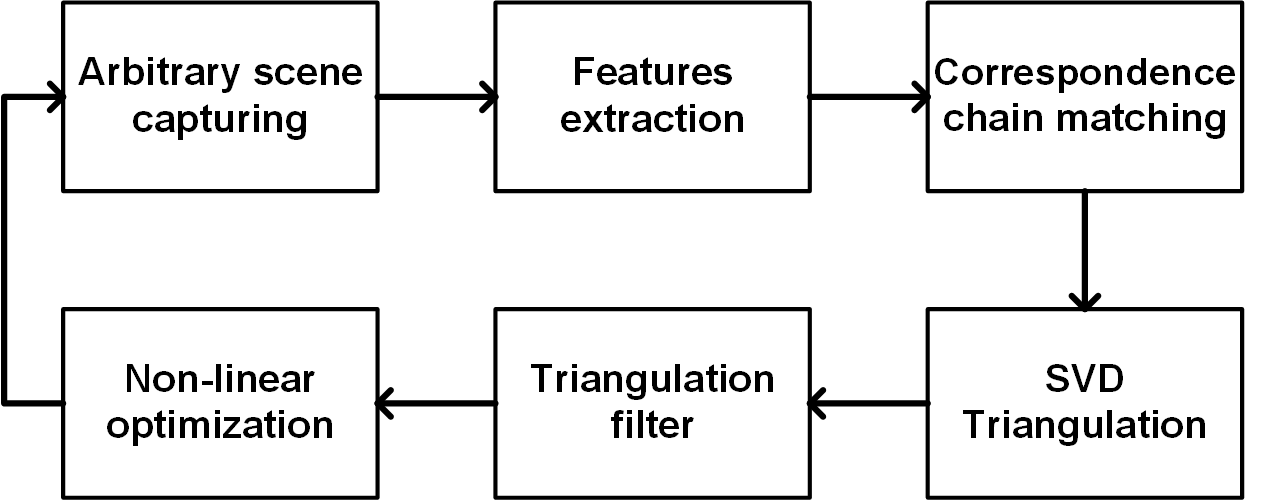}
    \caption{A pipeline of auto-refinement algorithm}
    \label{fig:autoref}
\end{figure}
An auto-refinement algorithm uses the previously found reference calibration (Section \ref{calibration}) and tries to estimate, how the calibration parameters need to be compensated for the current configuration of the multi-view camera. 
It is needed in cases when the placement of views was changed during the camera exploitation, \eg camera is mounted on the car and it is a subject of shakes and other mechanical influences. 
A pipeline of this method is presented in Fig. \ref{fig:autoref}.
\subsection{Correspondence Chain Matching} \label{ccm}
\begin{figure}[]
    \centering
    \includegraphics[width=7cm]{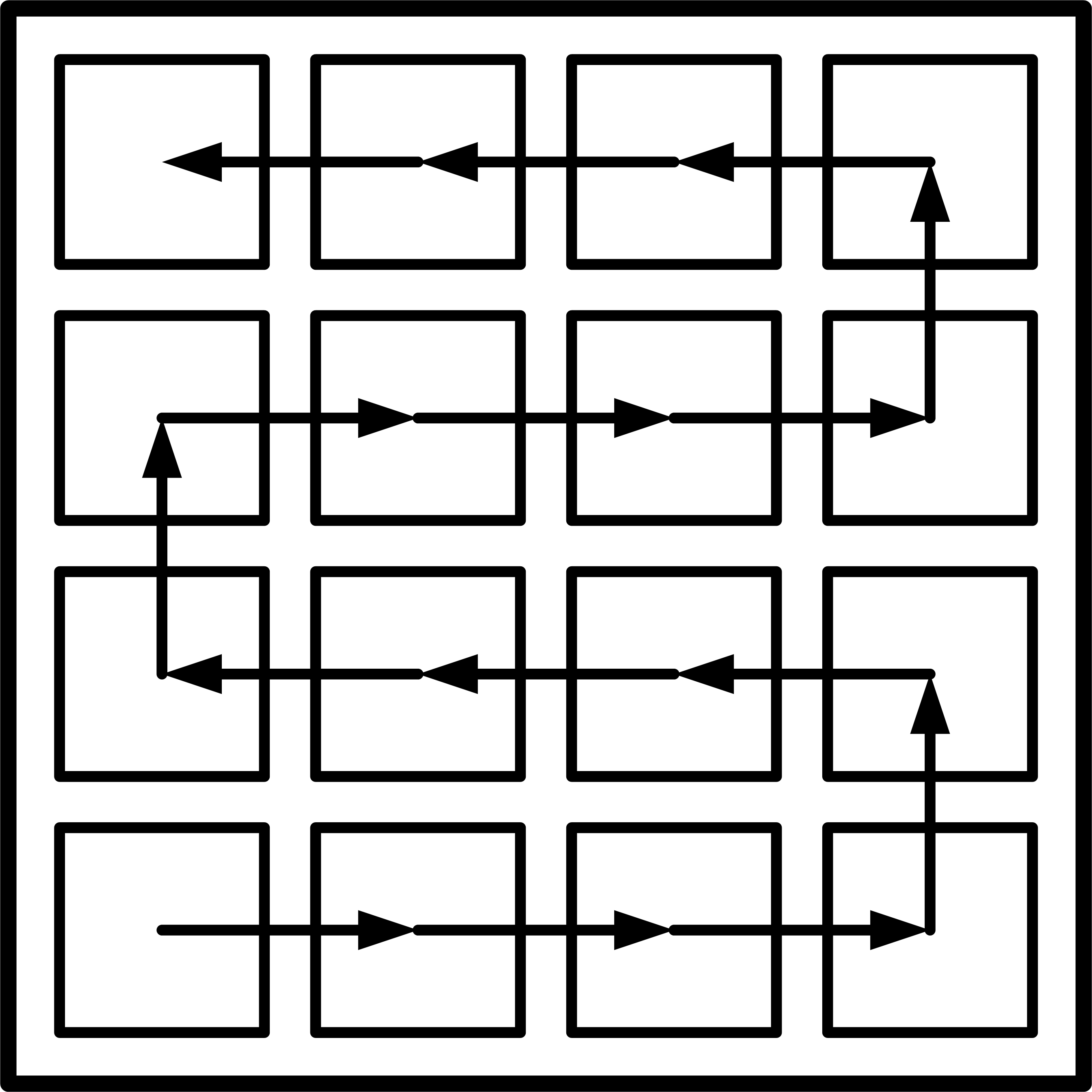}
    \caption{Visualization of views traversing in chain manner}
    \label{fig:ccm}
\end{figure}

The important criteria for the selection of correspondences detection algorithm were the robustness of features in real-world images and the running time of the algorithm. 
Among existing methods, the combination of "Good features to track" method for features extraction \cite{shi1994good} with Kanade-Lucas-Tomasi (KLT) feature tracker \cite{tomasi1991detection} were chosen. 
Exploiting the multi-view nature of the light field, in particular the fact that the projection of 3D points from the scene can be seen in all views, the determined features are tracked between different views in a chain manner. 
Correspondences from the reference view are verified in the neighboring view on the same axis, tracked ones are searched in the next view, and on the last view in one row features are matched in the upper one. 
This principle is demonstrated in Fig. \ref{fig:ccm}. 
Such a method is needed for reducing the number of possibly mismatched correspondences, which may occur especially in the small-baseline multi-view systems, like the camera used in Section \ref{experiments}. 
This method also helps for the verification of the correspondence inaccuracies, since the very strong matches have to be preserved in all views. 
Additional filtration, based on the estimation of fundamental matrix between correspondences in adjacent views, and exclusion of non-matched features is applied. 
Having a fundamental matrix $F$, estimated by \eg Random Sample Consensus method \cite{fischler1981random}, the points from neighboring view $x_{1}$ and $x_{2}$ are checked by the value of $x_{2}^TFx_{1}$ being lower than a certain threshold. 
There's a small number of features, which were extracted more than once from the image. 
To remove the possible influence of such correspondences, we preliminary check the result of the feature detection algorithm by searching of the nearby correspondences using the Euclidean distance between points. 
By empirically setting a certain threshold for these distances we can efficiently filter out the closely placed features.
For our experiments it was set to $\sqrt{2}$.
\subsection{Triangulation Filter} \label{tf}
Previously described methods of filtration can eliminate a big amount of wrongly estimated correspondences. 
However, some false matches, especially the ones placed close or on the textureless areas, can survive these checks. 
To eliminate such mismatches we propose a filter, based on the re-projection of triangulated points. 
The filtered points are triangulated based on original intrinsic values $K_{i}$ and rotation and translation vectors $R_{Oi}, t_{Oi}$, $i = 1..N$. 
A projection matrix $P^{O}_{i}$ is composed as $K_{i}[R^{O}_{i}|t^{O}_{i}]$. 
For every correspondence $m_{i} = [x_{i}, y_{i}]$, matched in every view out of $N$, by taking a vector $P_{oi}^{rT}$ for every row of the corresponding projection matrix the matrix $A$ is composed as follows \cite{hartley2003multiple}:
\begin{align}
A = 
\begin{bmatrix}
x_{1}P_{O1}^{3T} - P_{O1}^{1T}\\\\
y_{1}P_{O1}^{3T} - P_{O1}^{2T}\\
\vdots\\
x_{N}P_{ON}^{3T} - P_{ON}^{1T}\\\\
y_{N}P_{ON}^{3T} - P_{ON}^{2T}\\
\end{bmatrix}
\end{align}
Singular Value Decomposition is applied to this matrix, the triangulated points are extracted from the smallest singular value of $A$. 
These points are used afterward for the optimization result in Section \ref{ba}. 
Using the $P_{O}$ we project the triangulated points $M_{t}$ to 2D space and estimate the Euclidean distance between original and projected points:
\begin{equation}
\begin{gathered}
m_{t} = PM_{t}\\
dist_{t} = \|m_{t} - m_{r}\|,
\end{gathered}
\end{equation}
where $m_{r}$ stands for an original correspondence from reference view. 
We find a median of all distances between each correspondence and its projection, and filter out all matches, for which the distance is bigger than the median value. 

\subsection{Bundle Adjustment} \label{ba}
The bundle adjustment problem \cite{triggs1999bundle} is solved for estimating the compensation intrinsic and extrinsic values. 
For every initially found point $x_{ik}$, $k=1..M$, where $M$ is a total number of points, in all views $i=1..N$ we try to minimize reprojection error as: 
\begin{align}
\sum_{i=1}^{N}\sum_{k=1}^{M}\|x_{ik} - Q(X_{k}, K_{i}, d_{i}, R_{i}, t_{i})\| 
\label{eq:reproj_error}
\end{align}
, where $Q()$ projects the 3D points to 2D plane, involving Eq. \ref{eq:xy1} -- \ref{eq:dist}.
\section{Experiments} \label{experiments}
\begin{figure}[]
    \centering
    \includegraphics[width=7cm]{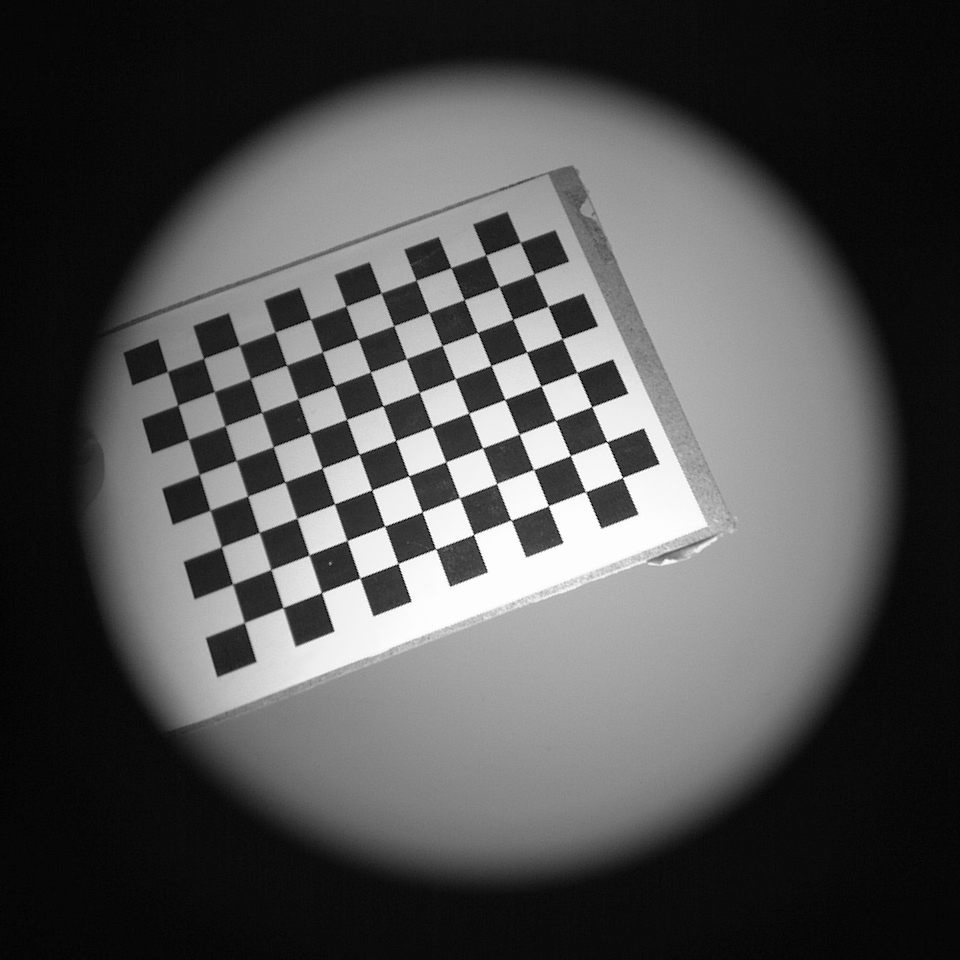}
    \caption{An example of checkerboard pattern, used for light field camera calibration}
    \label{fig:checkerboard}
\end{figure}

\subsection{Calibration algorithm} \label{rwlf}
\begin{figure}[]
    \centering
    \includegraphics[width=7.5cm]{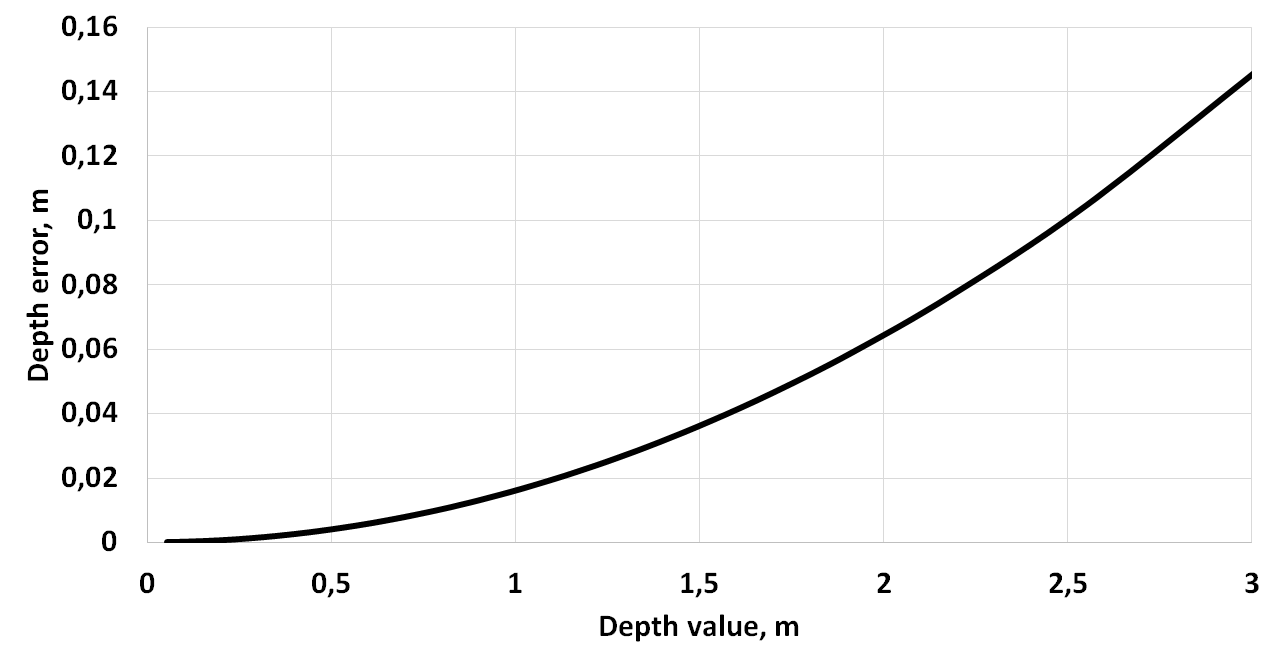}\\
    \caption{Depth error for $E_{PnP} = 0.246$ pixels}
    \label{fig:depth_error_2d}
\end{figure}

For the tests of pattern-based calibration algorithm we used a light field camera from \cite{anisimov2019a}.
It composed of 4x4 lenses, providing a 960x960 pixels RGB image per view, which afterward converted to grayscale for the sake of calculations simplicity.  
A 12x9 checkerboard pattern with 20x20 mm squares was used for the calibration. 
An example of the calibration pattern is presented on Fig. \ref{fig:checkerboard}.
We captured 40 scenes with this pattern for testing purposes.
To control the accuracy of the estimated intrinsic matrix $K$ and distortion coefficients $d$, the reprojection error is computed as in Eq. \ref{eq:reproj_error} for every light field view. 
Similar computations are done during the estimation of the extrinsic values. 
For the test dataset the average reprojection errors for monocular calibration and PnP problem were $E_{mono} = 0.159$ and $E_{PnP} = 0.246$ pixels respectively.
A comparison of our method was done with the method of Xu \etal \cite{xu2015camera}. 
For the same test dataset we have obtained reprojection errors of $E_{mono} = 0.153$ and $E_{PnP} = 0.156$ pixels.
We can state that optimization of all parameters together for the precisely estimated points make sense in terms of accuracy.
However, the potential drawback is related to higher running time of the joined optimization.
To verify influence of the reprojection error to the actual depth reconstruction we found a depth error with common focal length values $f = f_{x} = f_{y} = 850$ pix and baseline between the two most distant views on the same axis $b = 0.018 $ m as:
\begin{align}
\label{eq:depth_error}
\Delta Z = \frac{fb}{d-E_{pnp}} + \frac{fb}{d+E_{pnp}},
\end{align}
, where $d$ is the disparity value, \ie displacement between pixels, which can be further converted to actual depth value. 
This error is visualized on Fig. \ref{fig:depth_error_2d}.
On a target range of the test camera, which is 0.5-2.0 m \cite{anisimov2019a}, the estimated reprojection introduces an inaccuracy in the amount of 0.004-0.065 m, which is lower then the depth accuracy on this specific distance range.

\subsection{Auto-refinement algorithm}
\begin{figure}[]
    \centering
    \includegraphics[width=7cm]{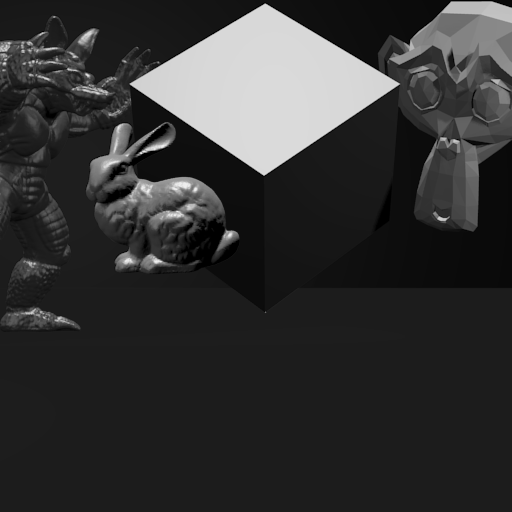} 
    \caption{An example of synthetic scene for verifying the auto-refinement algorithm}
    \label{fig:datasets}
\end{figure}
\begin{figure}[]
    \centering
    \includegraphics[width=7.5cm]{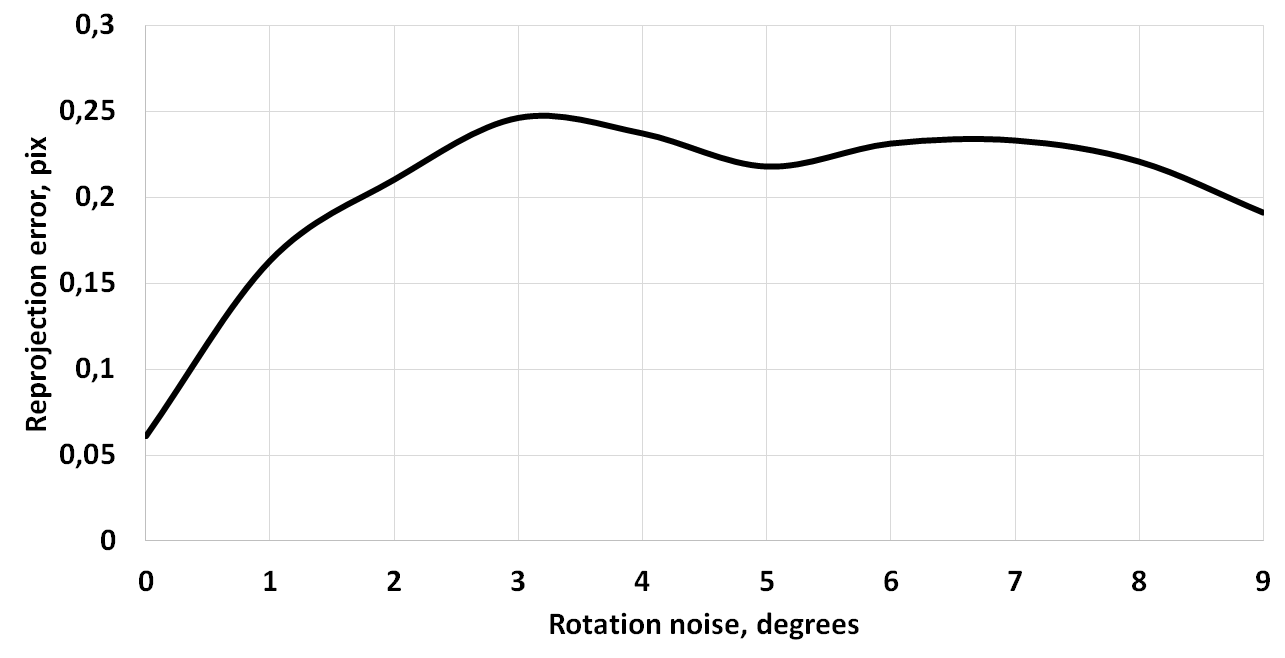}\\
    \includegraphics[width=7.5cm]{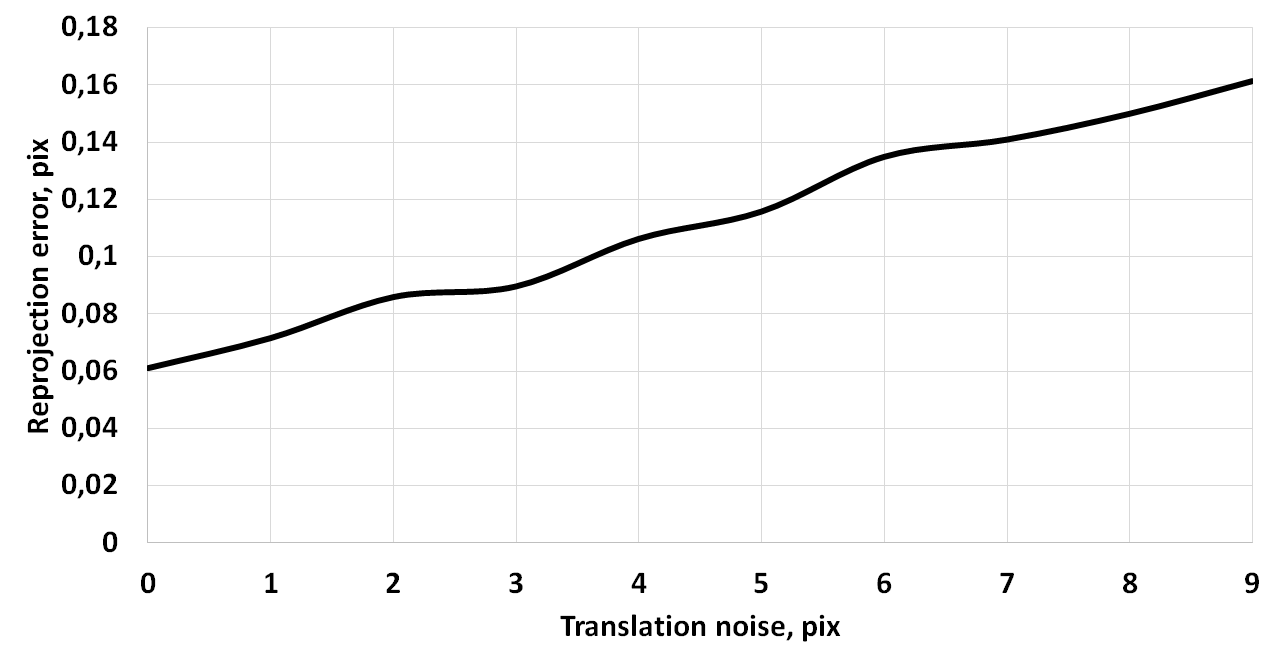}\\    
    \caption{Reprojection error of auto-refinement algorithm dependently of the applied noise for synthetic images}
    \label{fig:avg_synth}
\end{figure}
\begin{figure}[]
    \centering
    \includegraphics[width=7.5cm]{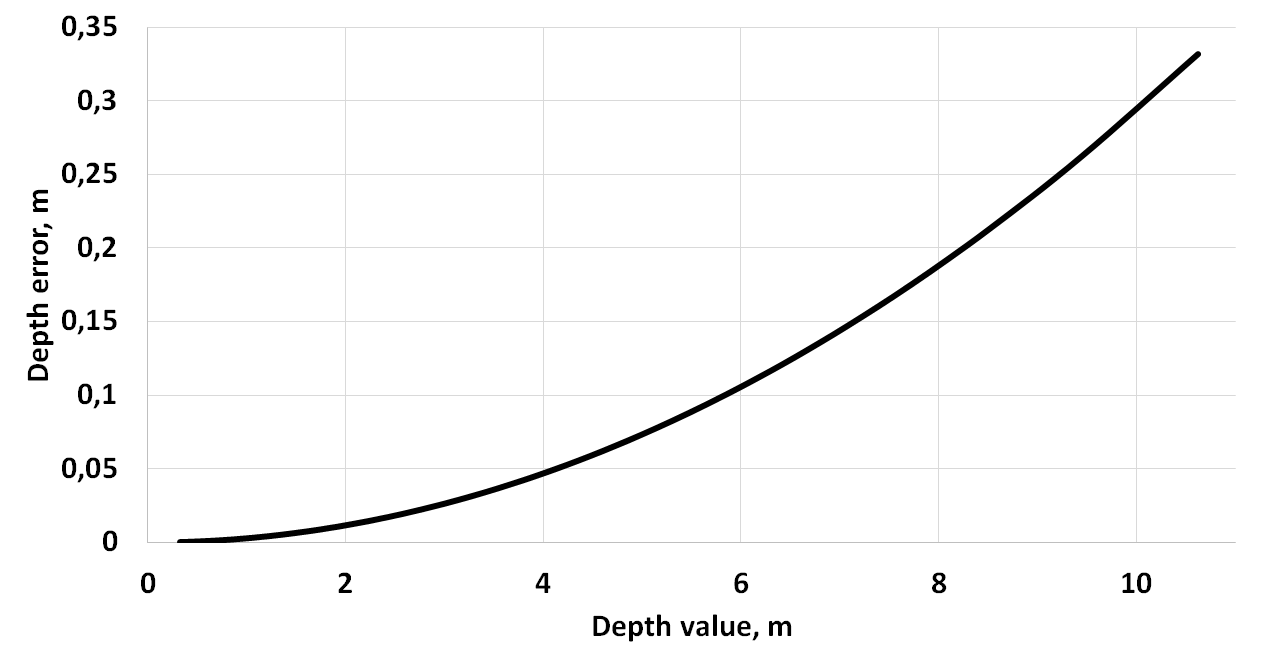}\\
    \caption{Auto-refinement depth error}
    \label{fig:depth_error_2d_autorefin}
\end{figure}

A dataset of synthetic light fields was used for verification of auto-refinement stability. 
The rectified and undistorted images with known intrinsic parameters were generated using Blender and the script from 4-dimensional Light Field Benchmark \cite{4DLFB} \cite{honauer2016dataset}. 
5x5 light fields of size 512x512 pixels per view with a baseline of 100 mm between adjacent views were generated from a simple scene with different overlapping objects, as demonstrated in Fig. \ref{fig:datasets}. 
A sequence of ten images was used for the tests. 
In average, 1/10 of originally detected points were filtered by correspondence chain matching, out of which half of the points survived the triangulation-based filtration. 
Processing of one frame takes around one second.
The difference between original and refined intrinsic and extrinsic parameters was measured and considered as non-informative, since no significant difference between original and refined values was found. 
It can be explained by the quality of correspondences, which is in general good for the synthetic data. 
To check the accuracy of the auto-refinement algorithm, we applied rotation and translation noise to the captured frames and measured the average reprojection errors.
Results are demonstrated on Fig. \ref{fig:avg_synth}. 
In total both types of noise create acceptable level of reprojection error. 
In a similar to pattern-based calibration manner we have evaluated the depth error for maximum reprojection error from the rotation noise, result of which is presented on Fig. \ref{fig:depth_error_2d_autorefin}. 
For the noise simulation of rotation was applied only to Z-axis.
Rotations on X and Y axes are the subject of tangential distortion. 
It occurs when the lens array is not parallel to the camera sensor plane.
This type of distortion is assumed to be zero in the applied camera model.
\subsection{Discussion}
%

%
The pattern-based calibration method used gives adequate results in terms of the resulting reprojection error. 
Empirically, we have found that the overall calibration quality depends largely on the quality of the calibration target, especially its flatness. 
For the proper calibration we have come to the number of 25-30 pattern images in one sequence. 
All areas on the light field views should be covered with pattern images in various positions to ensure correct estimation of internal values. 
One of the assumptions of the algorithms was the similarity of the lens parameters for each view. 
For cases with a significant shift (in terms of rotation or translation) of at least one of the views over the others, averaging over extrinsic values cannot be used; it should be replaced by nonlinear methods. 
Experiments with the auto-refinement algorithm on synthetic images show that the reprojection error increases in proportion to the level of lens shift, while changes in their rotation affect only up to some point with a plateau thereafter.  
During the auto-refinement experiments, we noticed that optimizing all the parameters together leads to incorrect results. 
This was the motivation for dividing the optimization procedure into three steps applied to the same model. 
First, only 3D points are optimized, while intrinsic and extrinsic camera values are fixed. 
This condition is relaxed in the second part, where the intrinsic values of all views are optimized. 
Finally, we optimize all camera parameters together. 
All of the above steps are repeated with new captured scenes. 
It can be stopped either by using a certain number of iterations or by reaching the desired reprojection error below a threshold value. 
Several limitations of the auto-refinement algorithm were identified during the tests. 
It does not work well with repeating textures and with small distances between detected matches, \eg on keyboards. 
This problem can be solved either by applying additional filtering measures or by changing the match detection algorithm.
In addition, the coverage of the image area by the correspondence is important to obtain correct lens geometry, a similar requirement stands for pattern-based calibration. 
Additional correspondence distribution checks can be made to discard images without proper coverage. 
Because of this, the distortion coefficients cannot be correctly corrected by our auto-rectification method and remain locked during the optimization stage.
We have tested the auto-refinement algorithm with and without triangulation filter. 
Without filtering the results were totally incorrect, so they are not included in part of the experiments.

\subsection{Implementation}
The algorithm was implemented in C++, uses only a central processing unit (CPU), and requires additional libraries such as OpenCV \cite{opencv} for image-related routines and Ceres Solver \cite{ceres} for solving nonlinear optimization problems. 
The algorithm was tested in Windows 7 with Intel Xeon CPU E3-1245 V2, additional tests were done on the NVIDIA Jetson TX2 and AGX Xavier platforms. 
Memory requirements directly depend on the size of the provided light field. 
\section{Conclusion} \label{conclusion}
In this paper we presented the calibration and rectification pipeline for light field cameras together with the method for calibration parameters refinement from the arbitrary scenes. 
Additional filtration measures like triangulation filter and two-dimensional technics were outlined. 
The algorithms were evaluated using synthetic and real-world data, the results of the experiments support the idea of algorithms' utilization for real-world light field calibration and its refinement. 
The presented calibration and auto-refinement principles can be further extended to other multi-view systems with preserved similarity of views alignment. 
To further our research we plan to overcome the mentioned limitations and conduct additional experiments on other similar multi-view systems.
For example, method from \cite{ddvernay2008straight} can be adapted to deal with distortions from arbitrary scene. 

\section{Acknowledgments}
This work was partially funded by the Federal Ministry of Education and Research (Germany) in the context of projects DAKARA (13N14318) and VIDETE (01IW18002). 
The authors are grateful to Jason Raphael Rambach and Torben Fetzer for the provided support.

\end{document}